\documentclass[runningheads]{llncs}
\usepackage[T1]{fontenc}
\usepackage{amsmath,amsfonts,bm}

\def\eqref#1{equation~\ref{#1}}

\def\1{\bm{1}}

\DeclareMathAlphabet{\mathsfit}{\encodingdefault}{\sfdefault}{m}{sl}
\SetMathAlphabet{\mathsfit}{bold}{\encodingdefault}{\sfdefault}{bx}{n}

\usepackage{epsfig}
\usepackage{graphicx}
\usepackage{multirow}
\usepackage{booktabs}
\usepackage{tabularx}
\usepackage{amsmath}
\usepackage{amssymb}
\usepackage{acro}
\usepackage{amsfonts}
\usepackage{algorithm2e}
\usepackage{xcolor}
\usepackage[colorlinks]{hyperref}

\begin{document}
\title{Low-Rank Continual Pyramid Vision Transformer: Incrementally Segment Whole-Body Organs in CT with Light-Weighted Adaptation}
\authorrunning{V. Zhu, Z. Ji et al.}
\titlerunning{Low-Rank Continual Pyramid Vision Transformer for Organ Segmentation}

\author{Vince Zhu\inst{1,2}, Zhanghexuan Ji\inst{1}, Dazhou Guo\inst{1}, Puyang Wang\inst{3}, Yingda Xia\inst{1}, Le Lu\inst{1}, Xianghua Ye\inst{4}, Wei Zhu\inst{2}, Dakai Jin\inst{1}}

\institute{DAMO Academy, Alibaba Group \and
State University of New York at Stony Brook, USA \and
Johns Hopkins University, USA \and
The First Affiliated Hospital Zhejiang University, Hangzhou, China \\
\email{\{zhanghexuan.ji, dakai.jin\}@alibaba-inc.com}}

\maketitle              
\begin{abstract}
Deep segmentation networks achieve high performance when trained on specific datasets. However, in clinical practice, it is often desirable that pretrained segmentation models can be dynamically extended to enable segmenting new organs without access to previous training datasets or without training from scratch. This would ensure a much more efficient model development and deployment paradigm accounting for the patient privacy and data storage issues. This clinically preferred process can be viewed as a continual semantic segmentation (CSS) problem. 
Previous CSS works would either experience catastrophic forgetting or lead to unaffordable memory costs as models expand. In this work, we propose a new continual whole-body organ segmentation model with light-weighted low-rank adaptation (LoRA). We first train and freeze a pyramid vision transformer (PVT) base segmentation model on the initial task, then continually add light-weighted trainable LoRA parameters to the frozen model for each new learning task. Through a holistically exploration of the architecture modification, we identify three most important layers (i.e., patch-embedding, 
multi-head attention and feed forward layers) that are critical in adapting to the new segmentation tasks, while retaining the majority of the pre-trained parameters fixed. Our proposed model continually segments new organs without catastrophic forgetting and meanwhile maintaining a low parameter increasing rate. Continually trained and tested on four datasets covering different body parts of a total of 121 organs, results show that our model achieves high segmentation accuracy, closely reaching the PVT and nnUNet upper bounds, and significantly outperforms other regularization-based CSS methods. When comparing to the leading architecture-based CSS method, our model has a substantial lower parameter increasing rate (16.7\% versus 96.7\%) while achieving comparable performance. 

\end{abstract}
\section{Introduction}

\begin{figure}[ht]
\centering
\includegraphics[width=1\columnwidth]{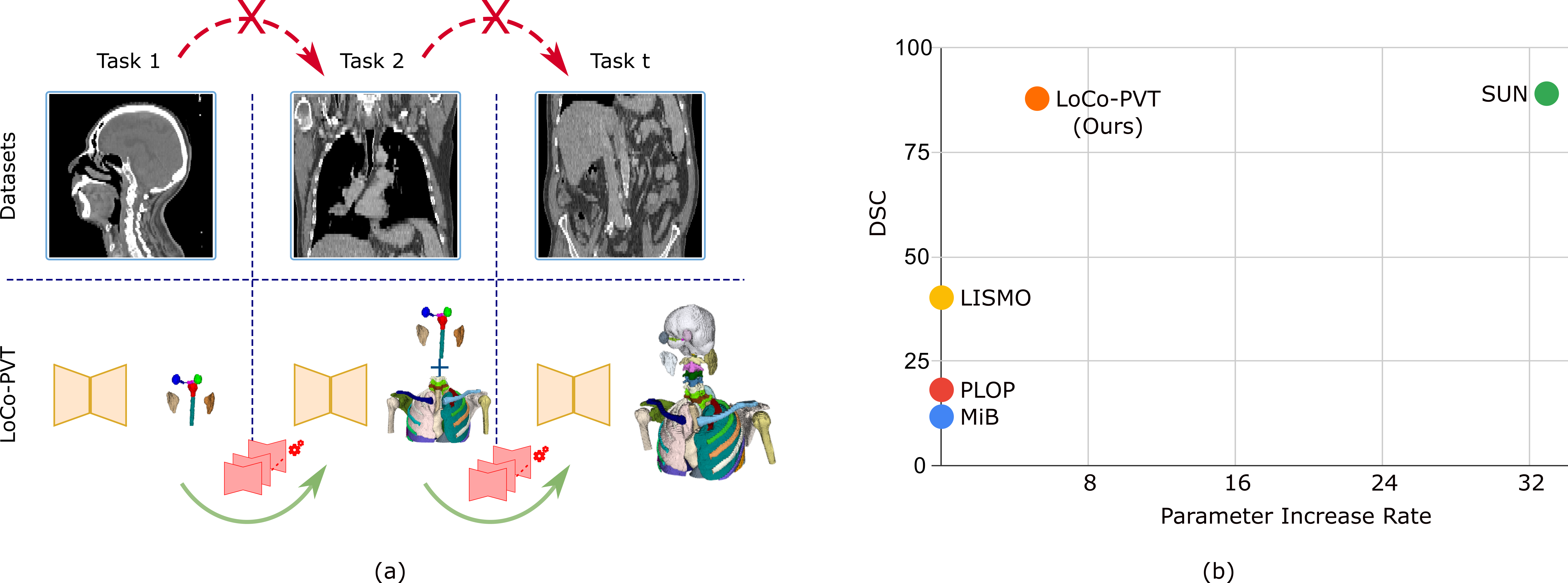}
\caption{Illustration of the continual multi-organ segmentation (a). At each continual learning step, only the previously trained model is available (green arrow). Previous datasets are not accessible. Illustration of the segmentation performance versus parameter increasing rate of continual multi-organ segmentation methods.} 
\label{fig:movtivation} 
\end{figure}

Multi-organ and tumor segmentation, one of the most essential medical image analysis tasks, has been widely studied in the literature~\cite{guo2020organ,shi2022deep,tang2019clinically}. With the fast development in deep learning segmentation techniques, deep segmentation networks trained on specific datasets achieve high performance comparable to those of medical experts~\cite{isensee2021nnu,jin2022towards,jin2021deeptarget,raju2020user,wang2023accurate,ye2022comprehensive}. However, current deep segmentation approaches are not capable of updating the trained models effectively when new segmentation classes are incrementally added, although in clinical practice it is desirable that pre-trained segmentation models can be dynamically extended to segment new organs without access to previous training datasets. Illustrated in Fig.~\ref{fig:movtivation}, this preferred process can be viewed as a continual semantic segmentation (CSS) problem, which is a non-trivial task because deep learning models suffer from catastrophic forgetting when fine-tuned directly on new dataset~\cite{kemker2018measuring,ma2023progressive,shen2024continual}.
CSS is emerging very recently in the natural image domain~\cite{cermelli2020modeling,douillard2021plop,zhang2022representation}, and the most common CSS approaches adopt the regularization constraint network training via knowledge distillation to reduce the forgetting of old knowledge while learning new classes. However, since entire network parameters are updated on the training of new classes, it is extremely difficult to achieve high performance on both old and new classes.
CSS has been rarely studied in medical imaging field~\cite{ji2023continual,liu2022learning,ozdemir2018learn,zhang2023continual}. Ozdemir et al. uses only 9 patients and 2 organ labels to develop a regularization-based continual segmentation model~\cite{ozdemir2018learn}. Liu et al. adopts the MiB loss~\cite{cermelli2020modeling} and prototype matching to continually segment a small number of 5 abdominal organs~\cite{liu2022learning}, and Zhang et al. utilizes the pseudo-labels and clip-embedded controller head to segment 13 abdominal organs~\cite{zhang2023continual}. Note that \cite{liu2022learning,zhang2023continual} both only focus on a limited organs in the abdomen CT, and when involving a large number of organs of various body parts, such as in whole-body CT scans, they suffer severe performance degradation (as demonstrated in our experiments later). Recently, a new architecture-based CSS method~\cite{ji2023continual} is proposed that avoids forgetting by freezing the CNN encoder after the initial task and sequentially adding separate decoder for each new task. Although it achieves high performance without forgetting, the method is less scalable because model parameters increase dramatically as new tasks are added (see Figure 1(b)). The completely frozen encoder also lacks of extensibility~\cite{ji2023continual}.

In this work, we aim to develop a new CSS method that avoids the catastrophic forgetting and meanwhile circumvent the model parameter explosion issue in~\cite{ji2023continual}. To achieve this goal, we take inspirations from two categories of recent technique advancements. First, 
vision transformer (ViT) is widely used in recent applications~\cite{dosovitskiy2020image,kirillov2023segment,wang2021pyramid}, and exhibits superiority in global feature extraction, self-supervised large model pretraining, and multi-modality learning as compared to the CNN-based models. In medical imaging, ViT-based models also demonstrate great potential for multi-organ segmentation task~\cite{tang2022self,xie2022unimiss,zhou2021nnformer}, since many of them exhibit comparable performance with the leading CNN-based models~\cite{isensee2021nnu}. 
Considering the capacity advantage of ViT models and its flexibility in being extended to diverse tasks, we envision that ViT-based architecture is suitable for CSS task.
Second, many recent parameter efficient fine-tuning (PEFT) methods are demonstrated to be effective when adapting the large scale pretrained language model to different downstream applications~\cite{ding2023parameter,houlsby2019parameter,hu2022lora}. For example, low-rank adaptation (LoRA)~\cite{hu2022lora} is one of the most popular and effective PEFT methods by freezing the pretrained model weights and injects trainable rank decomposition matrices into the linear or convolution layer of ViT. Hence, we assume that PEFT is capable of extending model's capacity to segment new organs with minor increased model parameters.

Motivated by the above observations, we propose a new architecture-based CSS method for continual whole-body organ segmentation using pyramid vision transformer with LoRA. We adopt the UniMISS~\cite{xie2022unimiss} pretrained 3D pyramid vision transformer (PVT) as backbone due to its large scale medical image pretraining and the leading performance on downstream segmentation tasks.  To circumvent the issue of catastrophic forgetting,  we introduce a subset of trainable parameters for each new task. Unlike previous methods that append a bulky decoder for each task~\cite{ji2023continual}, our approach utilizes LoRA on selected PVT layers to incrementally expand its capacity for segmenting new organs. Following the original LoRA configuration, a group of LoRA matrices are first injected to query \& value projection layers in multi-head attention to enhance the feature extraction. Furthermore, through a holistically exploration of the architecture modification, we inject LoRA matrices to the feed-forward network (FFN) to provide extra feature aggregation capability necessary for adapting to new unseen tasks. Additionally, we further extend LoRA matrices to 3D convolution in patch embedding layers of encoder and the last layer of decoder, which it critical to handle the large spacing variation in different medical segmentation tasks. Continually trained and tested on four datasets covering different body parts of a total of 121 organs, results show that our model achieves high segmentation accuracy, closely reaching the PVT~\cite{xie2022unimiss} and nnUNet~\cite{isensee2021nnu} upper bounds, and significantly outperforms other regularization or pseudo-label based CSS methods~\cite{cermelli2020modeling,douillard2021plop,zhang2023continual}.


\section{Method}


\begin{figure}[t]
\centering
\includegraphics[width=\columnwidth]{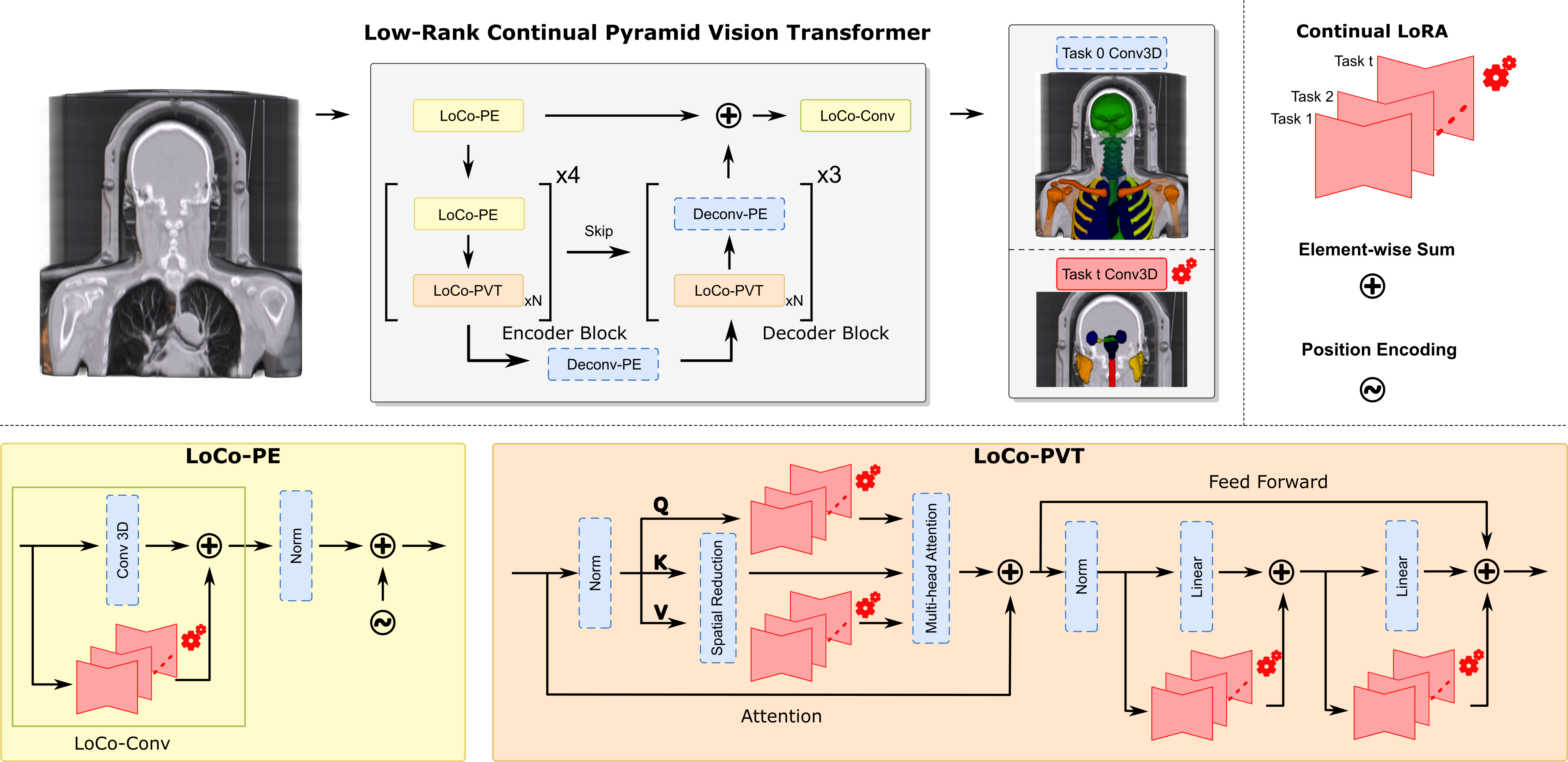}
\caption{Overall framework of the proposed low-rank continual pyramid vision transformer (LoCo-PVT) network for continual whole-body organ segmentation, which is composed of a stack of encoder and decoder blocks, where each block contains a patch embedding (PE) layer and multiple LoCo-PVT layers. 
Encoder PE layer (LoCo-PE) has a convolution layer with stride 2 for downsampling, while decoder PE layer (Deconv-PE) uses deconvolution layer for upsampling instead. Continual LoRA is added on linear layers for Q/V projection in multi-head attention and feed-forward network in LoCo-PVT, and is also added on convolution layers (LoCo-Conv) in LoCo-PE. 
The base network is frozen (colored in blue) after training the inital task 0. At each following continual learning step, a set of trainable LoRA parameters and a new segmentation output layer (colored in red) are added for new task adaptation.} 
\label{fig:loco_pvt} 
\end{figure}

\subsection{Problem Formulation}
Figure \ref{fig:loco_pvt} illustrates the proposed low-rank continual (LoCo) multi-organ segmentation framework. We adopt the UniMISS~\cite{xie2022unimiss} pretrained 3D PVT as backbone. Subsequently, the 3D PVT undergoes further training with the TotalSegmentator dataset~\cite{wasserthal2022totalsegmentator}. After this additional training, the PVT backbone, as depicted by the  blue dashed-line blocks in Figure~\ref{fig:loco_pvt}, is fixed throughout the subsequent training process. For the remaining tasks, let $D = \{D_1,\ldots,D_T\}$ represent the datasets sequence. The model is trained sequentially on each $D_t$ where $t\in\{1,\ldots,T\}$ and will not re-access $D_t$ after training is complete. Consider the $t^{th}$ dataset $D_t = \{X_k^t,Y_k^t\}_{k=1}^{N_t}$ that compromises $O_t$ organ classes, and assuming $(X^t$,$Y^t)$ denote all input images and the corresponding segmentation masks in $D_t$, the prediction map for voxel location $i$ and organ class $c_j$ is given by
\begin{align}
    \hat Y^t(i) = f\left(Y^t(i)=c_j\vert X^t; W_0, W_{L^t}, W_{S^t}\right), 
    \ \ \ 
    \hat{\mathbf Y} = \bigcup_{t=1}^T \hat Y^t ,
\end{align}
where $f$ denotes the transformer neural networks with frozen parameters $W_0$ trained on initial task and task-specific trainable LoRA parameters $W_{L^t}$ where $L^t = (A_l^t, B_l^t)_{l=1}^{M^t}$ denotes the pairs of low rank matrices for the $t^{th}$ dataset, and $W_{S^t}$ denotes the trainable task-specific Sigmoid output. The final prediction $\hat{\mathbf Y}$ is the union of all previous predictions with possible class overlapping. 



\subsection{LoCo-PVT: LoRA Continual Vision Transformer Layer}


Our LoCo-PVT framework inherits key advantages of both methods and is customized for continual multi-organ segmentation. The low-rank adaptations are enabled within every transformer block as well as the patch embedding modules of the encoder. For each dataset $D_t$, we associate a small set of trainable LoRA parameters $L^t = (A_l^t,B_l^t)_{l=1}^{M^t}$ where $(A_l^t,B_l^t)$ denotes LoRA matrices of the $l^{th}$ LoCo-PVT block and $M^t$ represents the total number of trainable LoCo-PVT blocks for the $t^{th}$ dataset.       

\subsubsection{LoCo-MHA \& LoCo-FFN}

For a pretrained weight matrix $W_0 \in \mathbb R^{d\times c}$ and the $t^{th}$ dataset, we constrain its update by representing the latter with a low-rank decomposition $W_0 + \Delta W^t = W_0 + B^tA^t$, where $B^t\in \mathbb R^{d\times r}, A^t\in \mathbb R^{r\times c}$, and the rank $r \ll\min(d,c)$. During training, $W_0$ is frozen and does not receive gradient updates, while $A^t,B^t$ contain trainable parameters. Both $W_0$ and $\Delta W$ are multiplied with the same input, and their respective output vectors are summed coordinate-wise. The forward pass of $h = W_0x$ can summarized as 
\begin{align}
    h = W_0x + \Delta W^t x
      = W_0x + B^tA^tx .
\end{align}
Each $A_t$ is initialized with random Gaussian and $B_t$ with a zero matrix, resulting in $\Delta W^t = B^tA^t$ being zero at the start of training. Then, $\Delta W^t x$ is scaled by $\frac{\alpha}{r}$ where $\alpha$ is a constant in $r$.

Although LoRA may be applied to any dense layer, Hu et al., \cite{hu2022lora} shows that applying it to queries and values of the MHA module yields the most significant performance gains. Therefore, we adopt a similar design choice in each LoCo-PVT block and uses a higher $r$ to accommodate for the greater complexity in learning from visual signals.

\subsection{LoCo-PE: LoRA Continual Patch Embedding Layer} The patch embedding modules project input patches into implicit embedding space of lower dimensions. Each encoder stage of LoCo-PVT is accompanied by a separate embedding module which extracts feature maps for various resolutions. Such spatial information are critical for training robust continual segmentation models across different datasets. Desirably, one should allocate trainable parameters to all convolutional projectors for each dataset. Compared to dense layers, it is observed that the inclusion of LoRA in convolutional layers resulted in a significant increase in number of parameters. To mitigate this issue, the application of convolutional LoRA is confined exclusively to the encoding PE layers, i.e., LoCo-PE layer, which incorporates LoRA-enabled 3D convolutions for projecting input patches to the embedding space in each encoder stage. To enhance feature aggregation and projection from all scales for the segmentation output, LoRA is also enabled in the second to the last convolution layer in decoder. 


In each 3D convolutional weight matrix $\delta_0\in \mathbb R^{d\times c\times k^3}$ of dataset $D_t$, a pair of matrices $(A^t,B^t)$ with the same rank $r$, where $B^t\in \mathbb R^{dk^2\times rk}, A^t\in \mathbb R^{rk\times ck}$, $k$ is kernel size. The forward pass of the convolutional operations are
\begin{align}
    \delta_0 \ast x + \Delta \delta \ast x
    = \delta_0 \ast x + (B^tA^t) \ast x , 
\end{align}
where $\delta_0$ is the frozen convolution weights from the pretrained $W_0$.
\\

{\noindent \bf Model Inference:} To merge the output probability maps from all learned tasks, we follows the body-part-aware output merging method from SUN~\cite{ji2023continual}, which pre-computes the average body part distribution map for each dataset, applies body-part regression over testing scans to eliminate the out-of-distribution body-part region from each task's prediction, then uses entropy-based ensemble to combine the prediction from all tasks. No task ID is required during inference. 

\section{Experiments and Results}

{\bf Datasets:} We evaluated the proposed model using the public dataset {\bf TotalSegmentator}~\cite{wasserthal2022totalsegmentator} (TotalSeg) as task 0 for base model training, which consists of 1204 CT scans of different body parts with 103 labeled anatomical structures (face label is removed). Similar to SUN~\cite{ji2023continual} dataset setting, we conduct continual segmentation on three in-house datasets which cover chest body part, head-neck body part and an esophageal dataset with tumor. {\bf Chest organ dataset} (CHO) contains 153 chest CT scans with 16 labeled chest organs, including 7 overlapping organs with TotalSeg and 9 new organs. {\bf Head-neck organ dataset} (HNO) includes 244 head \& neck CT scans with 9 new organs are annotated. The last {\bf esophageal dataset} (EsoTumor) contains 567 CT scans of esophagus with tumor, which is more challenging for esophagus segmentation. We use $80\%:20\%$ split for training and testing. At the final stage, a total of 121 organs are learned from all datasets. The median voxel resolutions are 
$1.5\times1.5\times1.5$mm, 
$1\times1\times2$mm, 
$0.7\times0.7\times5$mm, 
and $1\times1\times5$mm for TotalSeg, HNO, CHO and EsoTumor datasets, respectively. 

{\bf CSS Protocols, Baselines and Metrics:} In CSS experiments, the model is updated on a sequence of datasets. At each step, only the current dataset is used for training while all the previous datasets are not accessible. Following SUN~\cite{zhang2023continual} setting, two CSS orders are validated in order to demonstrate the robustness of the method. Order A: \textit{TotalSeg $\rightarrow$ CHO $\rightarrow$ HNO $\rightarrow$ EsoTumor}. Order B: \textit{TotalSeg $\rightarrow$ HNO $\rightarrow$ CHO $\rightarrow$ EsoTumor}, as shown in Table~\ref{tab:main}. We compare our method with 4 leading CSS works: 2 popular regularization-based baselines (MiB~\cite{cermelli2020modeling}, PLOP~\cite{douillard2021plop}), a regularization-based method in medical (LISMO~\cite{liu2022learning}) and an latest architecture-based method (SUN~\cite{ji2023continual}). All methods are implemented in nnUNet data preprocessing and augmentation framework. Our method uses PVT as backbone, while the other 4 methods are based on CNN. In this study, the upper bound of both PVT and nnUNet on each dataset are listed in Table~\ref{tab:main}. We report the final continual segmentation performance using the Dice similarity coefficient (DSC) and the $95\%$ Hausdorff distance (HD95). 

{\bf Implementation Details:} The 3D PVT base model is the same as UniMISS~\cite{xie2022unimiss} model-small, which contains 4 encoder blocks with $[2, 3, 4, 3]$ PVT layers each and 3 decoder blocks with [3, 4, 3] PVT layers each. The stride is 2 for the convolution in encoder PE and deconvolution in decoder PE for down-/up-sampling purpose. For LoRA setting, we set rank as 64, 16 for query/value layers and FFN in LoCo-PVT and 16 for convolution layers in LoCo-PE/LoCo-Conv; LoRA alpha is consistently set as half of corresponding rank. The encoder is initialized from Unimiss self-supervised pretrained parameters. Following Unimiss setting, batch size of 2 and patch size of $224\times224\times32$ are used for all datasets and the experiments. The ratio between the training and validation set is 4:1. All experiments are trained using AdamW and we set 6000 epochs for training base model on TotalSeg and 500 epochs for continual training steps. The initial learning rate for PVT is set as $1e^{-4}$ and weight decay as $3e^{-5}$. Models are trained on single NVIDIA A100 GPU. 


\newcolumntype{Y}{>{\centering\arraybackslash}X}

\begin{table}[t]
\centering
\caption{Final step results of benchmark CSS methods on two orders of the selected multi-organ datasets. Dataset names are followed by their class numbers. Mean DSC (\%, $\uparrow$) and HD95 (mm, $\downarrow$) are evaluated on each dataset as well as all classes (All). `PIR (\%, $\downarrow$)': parameter increasing rate of the final model (after three continual steps) compared to the size of base model trained on TotalSeg (initial step).} 
\label{tab:main}
\scriptsize
\begin{tabularx}{\textwidth}{l r *{10}{Y}}
\toprule
\multicolumn{1}{l|}{\multirow{2}{*}{\textbf{Methods}}} &
  \multicolumn{2}{c|}{\textbf{TotalSeg (103)}} &
  \multicolumn{2}{c|}{\textbf{CHO (16)}} &
  \multicolumn{2}{c|}{\textbf{HNO (9)}} &
  \multicolumn{2}{c|}{\textbf{EsoTumor (1)}} &
  \multicolumn{2}{c|}{\textbf{All (121)}} &
  \multicolumn{1}{c}{\multirow{2}{*}{\textbf{PIR}}} \\ \cmidrule(lr){2-11}
\multicolumn{1}{l|}{} &
  \textbf{DSC} &
  \multicolumn{1}{r|}{\textbf{HD95}} &
  \textbf{DSC} &
  \multicolumn{1}{r|}{\textbf{HD95}} &
  \textbf{DSC} &
  \multicolumn{1}{r|}{\textbf{HD95}} &
  \textbf{DSC} &
  \multicolumn{1}{r|}{\textbf{HD95}} &
  \textbf{DSC} &
  \multicolumn{1}{r|}{\textbf{HD95}} &
  \multicolumn{1}{c}{} \\ \midrule
\multicolumn{12}{c}{\textbf{Order A: TotalSeg $\rightarrow$ CHO $\rightarrow$ HNO $\rightarrow$ EsoTumor}} \\
\multicolumn{1}{l|}{\textbf{MiB}~\cite{cermelli2020modeling}} &
  9.18 &
  \multicolumn{1}{r|}{116.38} &
  28.55 &
  \multicolumn{1}{r|}{20.36} &
  9.21 &
  \multicolumn{1}{r|}{7.40} &
  87.35 &
  \multicolumn{1}{r|}{4.43} &
  12.19 &
  \multicolumn{1}{r|}{96.00} &
  \multicolumn{1}{r}{0} \\
\multicolumn{1}{l|}{\textbf{PLOP}~\cite{douillard2021plop}} &
  37.92 &
  \multicolumn{1}{r|}{53.49} &
  66.98 &
  \multicolumn{1}{r|}{19.60} &
  34.68 &
  \multicolumn{1}{r|}{15.55} &
  83.43 &
  \multicolumn{1}{r|}{5.97} &
  41.65 &
  \multicolumn{1}{r|}{46.27} &
  \multicolumn{1}{r}{0} \\
\multicolumn{1}{l|}{\textbf{LISMO}~\cite{liu2022learning}} &
  11.71 &
  \multicolumn{1}{r|}{137.65} &
  43.07 &
  \multicolumn{1}{r|}{29.07} &
  9.22 &
  \multicolumn{1}{r|}{12.93} &
  87.47 &
  \multicolumn{1}{r|}{4.45} &
  16.01 &
  \multicolumn{1}{r|}{114.45} &
  \multicolumn{1}{r}{0} \\
\multicolumn{1}{l|}{\textbf{SUN}~\cite{ji2023continual}} & 
  91.93 &
  \multicolumn{1}{r|}{3.44} &
  84.33 &
  \multicolumn{1}{r|}{5.20} &
  84.79 &
  \multicolumn{1}{r|}{2.67} &
  86.59 &
  \multicolumn{1}{r|}{5.15} &
  90.45 &
  \multicolumn{1}{r|}{3.62} &
  \multicolumn{1}{r}{96.7} \\
\multicolumn{1}{l|}{\textbf{Ours}} &
  91.07 &
  \multicolumn{1}{r|}{4.09} &
  81.51 &
  \multicolumn{1}{r|}{5.77} &
  82.90 &
  \multicolumn{1}{r|}{3.08} &
  84.56 &
  \multicolumn{1}{r|}{5.48} &
  89.26 &
  \multicolumn{1}{r|}{4.24} &
  \multicolumn{1}{r}{16.7} \\ \midrule
\multicolumn{12}{c}{\textbf{Order B: TotalSeg $\rightarrow$ HNO $\rightarrow$ CHO $\rightarrow$ EsoTumor}} \\
\multicolumn{1}{l|}{\textbf{MiB}~\cite{cermelli2020modeling}} &
  11.24 &
  \multicolumn{1}{r|}{145.78} &
  78.65 &
  \multicolumn{1}{r|}{7.23} &
  9.09 &
  \multicolumn{1}{r|}{24.83} &
  87.27 &
  \multicolumn{1}{r|}{4.36} &
  20.04 &
  \multicolumn{1}{r|}{119.06} &
  \multicolumn{1}{r}{0} \\
\multicolumn{1}{l|}{\textbf{PLOP}~\cite{douillard2021plop}} &
  31.58 &
  \multicolumn{1}{r|}{63.78} &
  79.47 &
  \multicolumn{1}{r|}{7.09} &
  22.78 &
  \multicolumn{1}{r|}{11.09} &
  83.19 &
  \multicolumn{1}{r|}{6.04} &
  37.31 &
  \multicolumn{1}{r|}{52.63} &
  \multicolumn{1}{r}{0} \\
\multicolumn{1}{l|}{\textbf{LISMO}~\cite{liu2022learning}} &
  15.01 &
  \multicolumn{1}{r|}{90.54} &
  79.49 &
  \multicolumn{1}{r|}{7.36} &
  8.93 &
  \multicolumn{1}{r|}{9.13} &
  87.32 &
  \multicolumn{1}{r|}{4.31} &
  23.14 &
  \multicolumn{1}{r|}{73.88} &
  \multicolumn{1}{r}{0} \\
\multicolumn{1}{l|}{\textbf{SUN}~\cite{ji2023continual}} &
  91.93 &
  \multicolumn{1}{r|}{3.44} &
  84.33 &
  \multicolumn{1}{r|}{5.20} &
  84.79 &
  \multicolumn{1}{r|}{2.67} &
  86.59 &
  \multicolumn{1}{r|}{5.15} &
  90.45 &
  \multicolumn{1}{r|}{3.62} &
  \multicolumn{1}{r}{96.7} \\
\multicolumn{1}{l|}{\textbf{Ours}} &
  91.07 &
  \multicolumn{1}{r|}{4.09} &
  81.51 &
  \multicolumn{1}{r|}{5.77} &
  82.90 &
  \multicolumn{1}{r|}{3.08} &
  84.56 &
  \multicolumn{1}{r|}{5.48} &
  89.26 &
  \multicolumn{1}{r|}{4.24} &
  \multicolumn{1}{r}{16.7} \\ \midrule\midrule
\multicolumn{12}{c}{\textbf{Single Task Upperbound}} \\
\multicolumn{1}{l|}{\textbf{PVT}~\cite{xie2022unimiss}} &
  91.07 &
  \multicolumn{1}{r|}{4.09} &
  83.56 &
  \multicolumn{1}{r|}{5.25} &
  84.67 &
  \multicolumn{1}{r|}{2.70} &
  87.02 &
  \multicolumn{1}{r|}{5.16} &
  89.66 &
  \multicolumn{1}{r|}{4.15} &
  \multicolumn{1}{r}{300} \\
\multicolumn{1}{l|}{\textbf{nnUNet}~\cite{isensee2021nnu}} &
  91.93 &
  \multicolumn{1}{r|}{3.44} &
  84.48 &
  \multicolumn{1}{r|}{5.14} &
  84.95 &
  \multicolumn{1}{r|}{2.66} &
  87.62 &
  \multicolumn{1}{r|}{4.39} &
  90.49 &
  \multicolumn{1}{r|}{3.60} &
  \multicolumn{1}{r}{300} \\ \bottomrule
\end{tabularx}%
\end{table} 

{\bf Comparisons to Other State-of-the-art Approaches:} 
The final continual segmentation results on two orders and single task upper bounds are shown in Table~\ref{tab:main}. On the previously learned three datasets and all organs, our method significantly outperforms 3 regularization-based methods (MiB~\cite{cermelli2020modeling}, PLOP~\cite{douillard2021plop}, LISMO~\cite{liu2022learning}) in both orders, where the severe catastrophic forgetting of these methods could be caused by large domain gap between different body parts. On the other hand, our method and SUN are both architecture-based methods hence have no forgetting over past tasks and are order invariant when base training dataset is the same (TotalSeg). Although SUN has a slightly higher mean Dice of 90.45\% than our 89.26\% on all organs, our parameter increasing rate is significant lower than SUN (16.7\% vs. 96.7\% on 3 continual tasks), since SUN adds an entire decoder for each new task while our method adds light-weighted low-rank adaptors in selected layers, which only increases 5.56\% per task. Note that, there is also a small gap between nnUNet upper bound 90.49\% and PVT upper bound 89.66\% on all organs, which shows a potential capability difference between nnUNet and PVT and might be the cause of the tiny performance gap between SUN (nnUNet-based) and LoCo-PVT (1.19\%). Our proposed method closely reaches the PVT upper bound on all organs with a marginal 0.4\% drop in DSC and a 0.9mm increase in HD95, which demonstrates the efficiency and effectiveness of continual LoRA with a well-trained frozen PVT. 

{\bf Ablation Study:} To demonstrate the importance of each continual LoRA components in the proposed LoCo-PVT network, we also conduct two ablation studies using one-step continual segmentation from TotalSeg to CHO, shown in Table~\ref{tab:abl_lora}. In the left table, various LoRA combinations are evaluated over three PVT components, including query \& value projection layer in multi-head attention (Attn-QV), feed-forward network in transformer layer (FFN) and 3D convolution in patch embedding layer (PE-Conv). Compared to `Base' setting, adding extra LoRA to FFN increases the CHO segmentation performance by 3.46\%, from 76.38\% to 79.84\%; adding extra LoRA to PE-Conv results in 80.36\%, which gains 3.98\%; our full LoCo design with LoRA in all three components further boosts the performance to 81.51\% and reduces HD95 from 7.96mm to 5.77mm. This ablation result shows that it is effective and essential to add LoRA in both FFN, which provides extra ability to project and ensemble new features from attention, and PE-Conv, which makes adaptation or localization on different patch resolution. In the right table, we further study the effect of adding LoRA in either encoder or decoder. In `encoder only' setting with frozen decoder, the mean DSC drops to 76.89\% and HD95 rises to 7.88mm; Similarly, in `decoder only' setting with frozen encoder, the mean DSC reduces to 77.21\% and HD95 increases to 7.52mm. The results shows that LoRA in both encoder (feature extraction) and decoder (organ localization) helps enhancing the adaptation ability of the network on new tasks and works equally important for LoCo-PVT network to get comparable performance with the upper bound. This ablation study validates the necessity of each component in our proposed network. 

\newcolumntype{Y}{>{\centering\arraybackslash}X}
\newcolumntype{P}[1]{>{\centering\arraybackslash}p{#1}}

\begin{table}[t]
\centering
\caption{One-step continual segmentation from TotalSeg to CHO trained with different LoRA ablation settings. LoRA ranks for Attn-QV, FFN and PE-Conv are set as 64, 16, 16, separately.}
\label{tab:abl_lora}
\parbox{.66\textwidth}{
\centering
\scriptsize
\begin{tabularx}{.65\textwidth}{l | P{13mm} P{8mm} P{13mm} | r r}
\toprule
\textbf{LoRA Settings} & \textbf{Attn-QV} & \textbf{FFN} & \textbf{PE-Conv} & \textbf{DSC} & \textbf{HD95} \\
\midrule
\textbf{Base}            & \checkmark &            &            & 76.38 & 7.96 \\
\textbf{Base w. FFN}     & \checkmark & \checkmark &            & 79.84 & 7.07 \\
\textbf{Base w. PE-Conv} & \checkmark &            & \checkmark & 80.36 & 6.30 \\
\textbf{Ours (Full)}     & \checkmark & \checkmark & \checkmark & 81.51 & 5.77 \\
\bottomrule
\end{tabularx}
}
\parbox{.32\textwidth}{
\centering
\scriptsize
\begin{tabularx}{0.31\textwidth}{l | r r }
\toprule
\textbf{LoRA Settings} & \textbf{DSC} & \textbf{HD95} \\
\midrule
\\
\textbf{Encoder only}  & 76.89 & 7.88 \\
\textbf{Decoder only}  & 77.21 & 7.52 \\
\textbf{Ours (Full)}   & 81.51 & 5.77 \\
\bottomrule
\end{tabularx}
}
\end{table} 

\section{Conclusion}
In this paper, we propose a new LoCo-PVT framework which combines LoRA with ViT for continual whole-body organ segmentation. We train and freeze a PVT base model on the initial task, then continually add light-weighted trainable LoRA parameters to the frozen base model, which avoids catastrophic forgetting and adapts the model to new tasks while maintaining a low parameter increasing rate. 
Our method achieves very high accuracy on 
four datasets covering different body parts, 
closely reaching the PVT upper bound, and outperforms other regularization-based methods. When comparing to leading architecture-based CSS method, our model exhibits a significantly lower parameter increase rate while achieving comparable performance. This efficiency highlights the effectiveness of our approach in optimizing resource use without compromising on the quality of organ segmentation. Future works include extending the LoCo-PVT to multi-modality datasets and other light-weighted ViT adaptation methods. 

\begin{credits}
\subsubsection{\discintname}
The authors have no competing interests to declare that are relevant to the content of this article.
\end{credits}

%
%
%
\bibliographystyle{splncs04}
\bibliography{Paper-0900.bib}

\end{document}